\documentclass{article}

\newif\ifarxiv\arxivtrue
\ifdefined\ARXIV\arxivtrue\fi

\usepackage{iclr2027_conference,times}
\ifarxiv\iclrfinalcopy\fi

\newif\ifanon\anontrue
\ifarxiv\anonfalse\fi

\usepackage[utf8]{inputenc}
\usepackage[T1]{fontenc}
\usepackage{hyperref}
\hypersetup{colorlinks=true, citecolor=blue, linkcolor=blue, urlcolor=blue}
\usepackage{url}
\usepackage{booktabs}
\usepackage{amsfonts}
\usepackage{amsmath}
\usepackage{amssymb}
\usepackage{microtype}
\usepackage{xcolor}
\usepackage{colortbl}
\usepackage{graphicx}
\usepackage{float}
\usepackage{array}
\usepackage{caption}
\usepackage{subcaption}

\newcolumntype{L}[1]{>{\raggedright\arraybackslash}p{#1}}

\newcommand\blfootnote[1]{%
  \begingroup
  \renewcommand\thefootnote{}\footnote{#1}%
  \addtocounter{footnote}{-1}%
  \endgroup
}

\title{Leak It: Per-Document Extraction Beyond\\
Aggregate Membership Inference}

\author{%
  Victor Maricato \\
  Karolinska Institutet, Sweden \\
  \texttt{victor.maricato.oliveira@stud.ki.se} \\
}

\begin{document}

\maketitle
\ifarxiv\lhead{Preprint.}\fi

\blfootnote{%
  \ifanon
    Code and audit tool: included with this submission as anonymised supplementary material.
  \else
    Code and audit tool: \url{https://github.com/victormaricato/leakit}
  \fi}

\begin{abstract}
Membership inference (MIA) on language models is usually summarised by an aggregate ROC-AUC, but recent work shows these evaluations are confounded: model-free \emph{blind} baselines separate members from non-members from surface text alone \citep{das2024blind}. Building on probabilistic discoverable extraction \citep{hayes2025probabilistic}, we study black-box, sampling-based training-data leakage using $N$ independent samples from $p_\theta(\cdot \mid x)$. We place mean overlap, extreme-value overlap, and self-concentration on a common functional-estimation footing. Under this view we confirm and extend the blind-baseline critique into the sampling regime: on WikiMIA a blind bag-of-words classifier reaches AUC $0.97$ (TPR@$5\%$FPR $0.90$) while sampling statistics add nothing, and on an IID Pile split (MIMIR) neither self-concentration nor gold-continuation recovery significantly exceeds a blind baseline in aggregate (incremental AUC $95\%$ CI includes zero). Aggregate metrics, however, hide the real harm. The same sampling \emph{verbatim-extracts} training data for a tail of documents that no blind attack can reach. On Pythia-6.9B, $16.6\%$ of $500$ Pile documents bearing a real identifier ($83$ documents; $21.3\%$ of those bearing an email address) have that exact identifier reproduced \emph{and} not reproduced under a mismatched-prefix control, so each leak is attributable to that specific document rather than to a globally common string. This is a per-document disclosure that aggregate AUC cannot see. The risk is uneven, and we report the decomposition rather than only the average: identifier leakage is ${\sim}3\times$ stronger in code than in prose, though prose stays clearly positive and grows with capacity ($4.0\%$ to $12.1\%$ from $410$M to $6.9$B), while recovery of \emph{arbitrary} held-out continuations is essentially confined to code ($+0.44$ member gap on GitHub vs ${\leq}{+}0.014$ on prose). We characterise the extraction regime: temperature and nucleus sampling have minor effect, a $16$-token prefix already suffices, and the measured sample-budget relationship corroborates prior probabilistic-extraction results. We detect no reduction from training on a deduplicated corpus. Our results argue that language-model privacy audits should report per-document extraction, not aggregate membership, and motivate differential privacy as the mitigation. We release \texttt{leakit}, a black-box tool implementing this probe and its control.
\end{abstract}

\section{Introduction}
\label{sec:intro}

Membership inference (MIA) asks whether a document was in a model's training set. It underlies privacy attacks, copyright audits, and unlearning verification. When a model is trained on a narrow, sensitive population, membership is itself a disclosure of the sensitive attribute: showing that a person's record was used to train a disease-specific model reveals that they have the condition \citep{knolle2026disparate}. For language models, MIA is almost always reported as an aggregate ROC-AUC over a benchmark of members and non-members. \citet{das2024blind} show this practice is unsound: on eight foundation-model benchmarks, model-free \emph{blind} classifiers that never query the model (bag-of-words, date detection) beat state-of-the-art MIAs, because members and non-members are drawn from distinguishable distributions. An aggregate AUC therefore certifies membership leakage only if it exceeds a blind baseline, and only on a benchmark whose members and non-members are identically distributed, for which \citet{das2024blind} recommend a corpus with an official train/test split such as the Pile; MIMIR \citep{duan2024mimir} provides exactly this. Likelihood-based attacks (LOSS \citep{yeom2018loss}, Min-K\% \citep{shi2024mink}, zlib ratio \citep{carlini2021extracting}, neighbourhood comparison \citep{mattern2023neighborhood}, ReCaLL \citep{xie2024recall}) further require token-level likelihoods, which closed APIs increasingly gate, and MoPe \citep{li2023mope} requires strictly more: white-box access to model parameters.

\paragraph{A probabilistic lens on sampling-based leakage.}
We ask what a black-box language model leaks about its training data through \emph{sampling} alone. Given a prefix $x$, we draw $N$ independent continuations $C_1, \dots, C_N \sim p_\theta(\cdot \mid x)$ and treat them as an empirical estimate of the model's output distribution. Sampling-based leakage signals are then \emph{functionals} of that distribution: the mean overlap between the samples and the document's held-out continuation (SaMIA; \citealp{kaneko2025samia}), its extreme value, and the self-concentration of the samples among themselves. This functional view (Sec.~\ref{sec:method}; illustrated in Figure~\ref{fig:pipeline}, Appendix~\ref{app:pipeline}) follows the probabilistic extraction of \citet{hayes2025probabilistic}, who define extraction as the probability that at least one of $n$ samples reproduces a target; it lets us put sampling attacks and extraction on the same probes, and is the analytic spine of our study.

\paragraph{Aggregate sampling-MIA is confounded too.}
Applying Das's discipline, we reproduce their blind baselines and compare them against every sampling functional. On WikiMIA a blind bag-of-words classifier reaches AUC $0.97$ (TPR@$5\%$FPR $0.90$) and sampling adds nothing; on the IID MIMIR split \citep{duan2024mimir} a temporal blind probe collapses to chance ($0.56$), confirming the split is clean, yet neither self-concentration nor gold-continuation recovery beats the residual blind baseline by a significant margin (incremental AUC $95\%$ CI includes $0$). As an \emph{aggregate} membership classifier, sampling does not add to what surface text already reveals.

\paragraph{But sampling verbatim-extracts a tail of training data.}
Aggregate AUC is the wrong lens: it can conceal per-sample instability, sharp variation across works, and a small set of highly exposed records \citep{hayes2025limits,cooper2026books,knolle2026disparate}. Extraction is exactly what a blind attack cannot reproduce. Over $500$ Pile documents each bearing a real email address or telephone number, black-box sampling on Pythia-6.9B reproduces the \emph{exact} identifier of $83$ of them ($16.6\%$) while a mismatched-prefix control shows the model does not emit that identifier generically, so the leak is attributable to that document. Restricted to email addresses the figure is $21.3\%$. This per-document disclosure is invisible to aggregate AUC and impossible for a text-only blind attack, and it is the concrete privacy harm MIA is meant to detect: if an individual's document was in training, sampling can reproduce their identifier. We further characterise the extraction regime: temperature and nucleus sampling have minor effect, a $16$-token prefix already suffices, and the sample budget needed scales inversely with prefix length.

\paragraph{Contributions.}
\begin{itemize}\itemsep1pt
  \item[(C1)] Building on the probabilistic extraction of \citet{hayes2025probabilistic}, a functional view that places mean-overlap, extreme-value, and self-concentration statistics on one footing as functionals of $p_\theta(\cdot\mid x)$ estimated from $N$ samples, which lets us compare sampling attacks and extraction on the same probes (Sec.~\ref{sec:method}).
  \item[(C2)] We extend the blind-baseline critique \citep{das2024blind} into the sampling regime: on an IID split, no sampling functional significantly beats a model-free baseline as an aggregate membership classifier (Sec.~\ref{sec:results}).
  \item[(C3)] A per-document criterion for identifier leakage: a mismatched-prefix control applied \emph{per document} rather than as a population rate, which attributes a leak to one document. This makes central the PII setting that \citet{hayes2025probabilistic} consider only cursorily (Sec.~\ref{sec:extraction}).
  \item[(C4)] We release \texttt{leakit}, which implements the per-document probe and its control, and argue that privacy audits should report per-document extraction rather than aggregate membership.
\end{itemize}

\section{Related Work}
\label{sec:related}

\paragraph{Sampling-based MIA (SaMIA).}
\citet{kaneko2025samia} introduced SaMIA, the first \emph{sampling-based} likelihood-independent membership-inference method, which scores a candidate using only sampled text. SaMIA splits a candidate $d$ into prefix $x$ and suffix $y$, samples $N$ continuations $C_i \sim p_\theta(\cdot\mid x)$, and uses $\bar{r}(d) = \tfrac{1}{N}\sum_i \text{ROUGE}_n(C_i, y)$ as the membership statistic (with an optional zlib reweighting). SaMIA evaluates on WikiMIA against logit baselines including LOSS and Min-K\%. We share SaMIA's no-logit threat model but go further: (i) we frame the problem as functional estimation on $p_\theta(\cdot\mid x)$, recovering SaMIA's $\bar{r}$ as one specific functional; (ii) we identify two complementary families (extreme-value and self-concentration) that respectively improve and replace SaMIA's mean; (iii) we evaluate on MIMIR \citep{duan2024mimir}, which SaMIA does not.

\paragraph{Other no-internals MIA families.}
Three recent lines reduce attacker access without becoming continuation-free. \emph{AttenMIA} \citep{zaree2026attenmia} trains a supervised classifier on cross-layer consistency and perturbation-shift features of the model's \emph{attention maps}, requiring white-box access to internals. \emph{PETAL} \citep{he2025petal} (USENIX Security 2025) is label-only but signals on per-token semantics of the candidate document, still needing $y$. \emph{SPV-MIA} \citep{fu2024spvmia} (NeurIPS 2024) prompts the target to bootstrap a calibration set for a reference model and runs likelihood-based MIA on the candidate. The continuation-free, black-box, sample-only setting (no internals, no logits, no $y$) is the strictly smaller surface this paper occupies.

\paragraph{Logit-based MIA.} The LOSS attack \citep{yeom2018loss} thresholds on document NLL. Subsequent variants reweight or recalibrate the per-token contribution (Min-K\%, \citealp{shi2024mink}; Min-K\%++, \citealp{zhang2024minkpp}), normalise against an external complexity baseline (zlib; \citealp{carlini2021extracting}), compare to a reference model (Ratio; \citealp{carlini2021extracting}) or a neighbourhood ensemble \citep{mattern2023neighborhood}, or condition on a non-member prefix (ReCaLL; \citealp{xie2024recall}). All require at least token-level likelihood access, and some require more: Ratio needs a second model and MoPe \citep{li2023mope} needs white-box parameters. MIMIR \citep{duan2024mimir} provides a unified head-to-head comparison of five widely used likelihood-based attacks.

\paragraph{Black-box extraction.} \citet{carlini2021extracting} demonstrated that sampling from GPT-2 with Internet-scraped prefixes can recover verbatim training-data sequences, and \citet{nasr2025scalable} scaled this to aligned production models. \citet{lukas2023analyzing} analyse PII leakage specifically, showing that scrubbing and differential privacy trade utility against measured PII exposure. These works establish that extraction is possible; we place it against the membership-inference literature, showing extraction persists in a tail precisely where aggregate membership fails a blind-baseline control.

\paragraph{Probabilistic extraction and per-example reporting.} \citet{hayes2025probabilistic} define $(n,p)$-discoverable extraction through the probability that at least one of $n$ samples exactly reproduces a target suffix. Their binary exact-match event is a special case of our extreme-value family, while our continuous overlap functionals also measure partial recovery. They show that extraction grows log-linearly in $n$ and derive the queries needed for a target probability, on Pythia among other families. Our $N^\ast$ values (Sec.~\ref{sec:ext_regime}) therefore corroborate that sample-budget relationship rather than establish it; we additionally sweep temperature and nucleus jointly where they focus on top-$k$. We extend this line in three directions: identifier-level PII extraction, a per-document mismatched-prefix control that tests whether the \emph{same} identifier is emitted from an unrelated context, and comparison against a model-free blind baseline. The broader case for per-example reporting also precedes ours: \citet{hayes2025limits} show that aggregate MIA metrics can conceal per-sample decision instability under training randomness, while \citet{cooper2026books} show that book memorisation varies sharply by model and work. Our distinct question is whether aggregate membership metrics conceal a tail of individually attributable identifier extraction.

\paragraph{Memorisation scaling.} \citet{carlini2022memorization} establish that LLM memorisation grows with model scale, with the number of times a training example is duplicated, and with prompt context length; \citet{tirumala2022memorization} show that larger models memorise faster. Our scale-ladder results are consistent with the scale trend: at fixed $N$, extreme-value sampling-MIA AUC increases monotonically across Pythia 410M $\to$ 6.9B.

\paragraph{Benchmarks.} WikiMIA \citep{shi2024mink} aligns Wikipedia events with each model's training cutoff. \citet{duan2024mimir} provide MIMIR, spanning seven Pile domains with additional $n$-gram deduplication between members and non-members ($13$-grams, ${\leq}80\%$ overlap). \citet{maini2024llm} show WikiMIA's pre-/post-cutoff shift inflates apparent AUC (Min-K\% reaches ${\approx}0.7$ there but ${\approx}0.5$ on IID Pile Wikipedia splits), and \citet{shi2024mink} document that detection correlates with text length, which MIMIR controls by bounding samples to $100$--$200$ words. \citet{das2024blind} generalise this into a blanket warning: model-free \emph{blind} classifiers (bag-of-words, date detection) beat state-of-the-art MIAs on eight foundation-model benchmarks (e.g.\ $94.7\%$ TPR@$5\%$FPR on WikiMIA), so a reported AUC evidences membership leakage only if it exceeds a blind baseline, and they recommend evaluating on a genuine train/test split such as the Pile. We adopt this discipline throughout: MIMIR is a Pile train-vs-test split, and we report a blind prefix classifier alongside every attack.

\paragraph{Individual-level and tail risk.} \citet{knolle2026disparate} show, for medical classifiers, that aggregate MIA AUC can sit near chance while a tail of individual records is highly vulnerable, that this vulnerable fraction grows with model capacity, and that partial-record access still succeeds for some individuals; they flag per-record extraction against \emph{generative} models as an open direction. That black-box generative setting, and the aggregate-versus-tail distinction, are central to how we report results.

\section{Sampling-MIA as Functional Estimation}
\label{sec:method}

\subsection{Framework}
\label{sec:method:framework}

Let $M$ be a target model with conditional density $p_\theta(\cdot \mid x)$. For a candidate document $d$, split as $x = (d_1,\dots,d_k)$ and $y = (d_{k+1},\dots,d_{k+\ell})$, a \emph{sampling-MIA statistic} is any functional $\mathcal{S}(\widehat{p}_N; y)$ of the empirical distribution $\widehat{p}_N$ formed by $N$ i.i.d. samples $C_i \sim p_\theta(\cdot\mid x)$, where the reference $y$ may be used or omitted. The classifier predicts ``member'' if $\mathcal{S} > \tau$; ROC sweeps $\tau$.

Three classes of functional are observable from the sample, with $\omega$ a text-overlap measure: \emph{mean overlap} $\mathcal{S}_{\text{mean}} = \tfrac{1}{N}\sum_i \omega(C_i, y)$, which is SaMIA's statistic \citep{kaneko2025samia}; \emph{extreme value} $\mathcal{S}_{\text{ext}} = T(\omega(C_1,y),\dots,\omega(C_N,y))$ for $T \in \{\max,\ \text{set-union recall}\}$; and \emph{self-concentration} $\mathcal{S}_{\text{self}} = \binom{N}{2}^{-1}\sum_{i<j}\omega(C_i, C_j)$, which never references $y$ and instead asks how concentrated $p_\theta(\cdot\mid x)$ is.

Mean and extreme value are different summaries of the same joint distribution $(\omega(C_1,y),\dots,\omega(C_N,y))$: when it is concentrated both are tight, but when membership produces a rare tail of near-verbatim continuations the mean smears that tail into noise while the maximum preserves it. This is why the extreme-value functional, not the mean, is the one that exposes extraction (Sec.~\ref{sec:extraction}).

\subsection{Sampling-MIA scorers used in this paper}
\label{sec:method:scorers}

All statistics lie in $[0,1]$ and are computed on the same $N$ completions $C_1,\dots,C_N \sim p_\theta(\cdot \mid x)$. The three we report in the main text are the extreme-value recovery of the held-out continuation, $s_{\text{ext-5gram}}(d) = |\mathrm{5gram}(y) \cap \bigcup_i \mathrm{5gram}(C_i)| \,/\, |\mathrm{5gram}(y)|$; SaMIA's mean overlap $s_{\text{ROUGE-}n}(d) = \tfrac{1}{N}\sum_i \mathrm{ROUGE}_n(C_i, y)$; and the $y$-free self-concentration $s_{\text{self}}(d) = \binom{N}{2}^{-1}\sum_{i<j}\mathrm{Jacc}(\phi(C_i), \phi(C_j))$, where $\phi$ extracts either character $k$-grams ($s_{\text{self-}k}$) or word tokens (the parameter-free $s_{\text{self-word}}$). Twelve further variants (max-LCS and max-Jaccard extremes, mean and variance forms, and a unique-string ratio) are defined with full per-configuration results in Appendix~\ref{app:full_grid}. A classifier predicts ``member'' if $s(d) > \tau$; ROC results sweep $\tau$.

\subsection{Logit-MIA baselines}

For comparability we compute four established logit-MIA scores on the same probe set:
LOSS \citep{yeom2018loss},
Zlib \citep{carlini2021extracting},
Min-K\% Prob ($K{=}20$; \citealp{shi2024mink}),
and Ratio against a smaller reference model (Pythia-160M for Pythia targets; Pythia-160M for OLMo as a cross-family check). All baselines see the full document $x \mathbin\Vert y$, not just $y$, matching MIMIR's convention.

\subsection{Probe set}

We use two probe sets. WikiMIA \citep{shi2024mink} at length splits 128 and 256 gives 332 documents ($190$ members, $142$ non-members; the buckets are not individually balanced), members drawn from pre-cutoff Wikipedia events and non-members from post-cutoff ones. We retain it only to show that it is separable from surface text alone (Sec.~\ref{sec:res_blind}) and draw no leakage conclusions from it. All leakage claims use MIMIR \citep{duan2024mimir}: members from the Pile \emph{train} split, non-members from the official Pile \emph{test} split, identically distributed by construction. We use only these splits, not MIMIR's temporally shifted Wikipedia and ArXiv variants. The default configuration is $k_{\text{prefix}}{=}64$, $\ell_{\text{target}}{=}64$; Sec.~\ref{sec:ext_regime} sweeps prefix length, temperature, and top-$p$.

\subsection{Models, decoding, statistics}

\textbf{Target models.} Pythia 410M, 1B, 2.8B, 6.9B \citep{biderman2023pythia} (trained on The Pile); OLMo-1B \citep{groeneveld2024olmo} (trained on Dolma).
\textbf{Reference model.} Pythia-160M, used for the Ratio baseline.
\textbf{Decoding.} Greedy off (sampling on); $T=1.0$; top-$p=1.0$; top-$k=0$; $N$ varied per experiment.
\textbf{Compute.} All runs executed on single commodity GPUs (T4 for targets $\leq 2.8$B parameters, A10G for 6.9B). Per-document checkpointing makes runs idempotent and resumable. Total compute for the headline experiments is ${<}50$ GPU-hours.
\textbf{Statistics.} Cluster-bootstrap confidence intervals (2{,}000 resamples, clustered by source). The N-sweep AUC curve sub-samples the $32$ saved completions per probe down to smaller $N$, single-draw (no replicate averaging), matching the headline runs.

\section{Aggregate membership inference is confounded}
\label{sec:results}

\subsection{Blind baselines beat sampling functionals}
\label{sec:res_blind}

\begin{figure}[!htbp]
  \centering
  \includegraphics[width=0.66\linewidth]{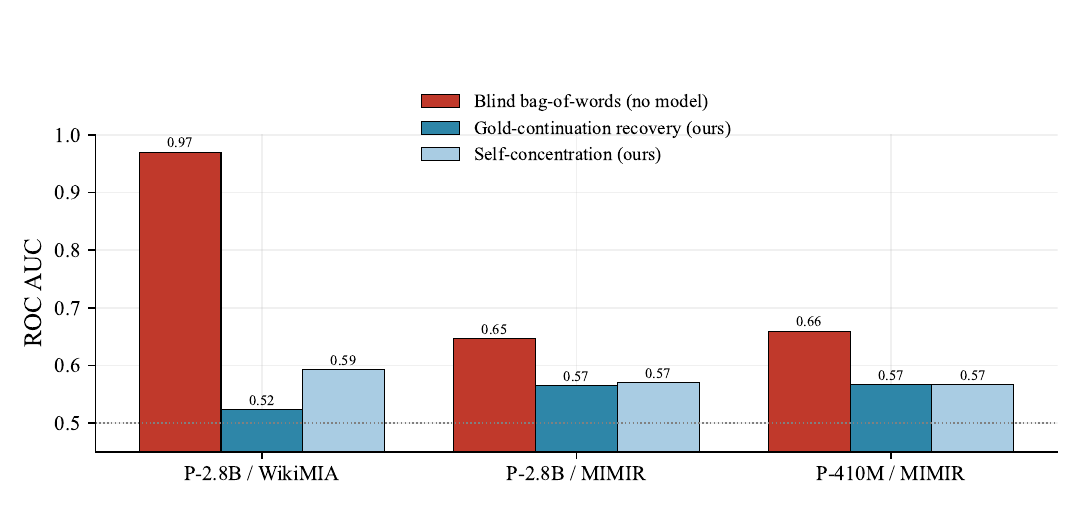}
  \caption{Does black-box sampling beat a model-free classifier? Aggregate ROC-AUC for a blind bag-of-words classifier (no model access) versus two sampling functionals. On WikiMIA the blind baseline is near-perfect; on the IID MIMIR split it retains a residual edge over both sampling statistics. Dotted line: chance (Pythia targets).}
  \label{fig:blind}
\end{figure}

Following \citet{das2024blind}, a sampling functional evidences membership leakage only if it exceeds a model-free \emph{blind} baseline on the same probes. We reproduce three blind attacks on the full document text: a date-detection threshold, a TF-IDF bag-of-words classifier (5-fold cross-validation), and a greedy rare-$n$-gram selector, and report ROC-AUC and TPR at low FPR (Figure~\ref{fig:blind}).

On \textbf{WikiMIA} the blind bag-of-words classifier reaches AUC $0.970$ and TPR@$5\%$FPR $0.90$, in line with the $94.7\%$ blind TPR@$5\%$FPR that \citet{das2024blind} report for the same benchmark; no sampling functional approaches it, and gold-continuation recovery is flat at AUC $0.52$. WikiMIA is therefore separable almost entirely from surface text, and any sampling AUC reported on it reflects that artefact rather than model leakage. On the IID \textbf{MIMIR} split \citep{duan2024mimir}, date-detection collapses to chance (AUC $0.556$), confirming the absence of a temporal shift; the residual blind bag-of-words classifier reaches AUC $0.646$ (TPR@$5\%$FPR $0.19$).

\subsection{No sampling functional adds significant signal}
\label{sec:res_null}

We test whether any sampling functional adds membership signal \emph{over} the blind baseline, the quantity that matters for Das's critique. On MIMIR, adding self-concentration to the blind classifier raises AUC by only $+0.001$ (macro); a best-case ensemble of every self-concentration variant (raw, capacity-contrast across the Pythia ladder, and scale-slope) reaches an incremental AUC of $+0.016$ with a cluster-bootstrap $95\%$ CI of $[-0.021, +0.046]$. Gold-continuation recovery, the strongest sampling signal, adds incremental AUC $+0.009$ ($95\%$ CI $[-0.009, +0.024]$) and incremental TPR@$5\%$FPR $+0.018$ ($95\%$ CI $[-0.005, +0.045]$). Every interval includes zero. As an \emph{aggregate} membership classifier on an IID benchmark, black-box sampling does not add to what a model-free reader of the text already recovers. This extends the blind-baseline critique of \citet{das2024blind} from logit-based MIA into the sampling regime.

\section{But sampling extracts a tail of training data}
\label{sec:extraction}

The aggregate view is the wrong lens: it averages over records, hiding a small set that are highly exposed \citep{knolle2026disparate}. We now measure the quantity a blind attack \emph{cannot} produce by construction, verbatim reproduction of held-out training content.

\subsection{Verbatim extraction of personally identifying information}
\label{sec:ext_pii}

To make the harm concrete we target real identifiers. From Pile-train documents (members: Pythia was trained on the Pile) we build a shared probe set of $500$ documents, $300$ containing an email address and $200$ a telephone number. For each we place the prefix immediately before the identifier, sample $N{=}32$ continuations at $T{=}1.0$, and ask whether any reproduces the \emph{exact} identifier; the threat model is an adversary holding text that precedes an identifier, the partial-record access \citet{knolle2026disparate} show still succeeds for some individuals. A raw rate would overstate the harm, since a model emits a globally common address in any context, so we add a \emph{context control}: the same query under a \emph{mismatched} prefix. We then take the conjunction \emph{per document} rather than subtracting two population rates, counting a document as leaked only if its identifier was reproduced \emph{and} was not reproduced under the mismatched prefix. This is what licenses a claim about individual documents.

On Pythia-6.9B, $83$ of the $500$ documents leak by this criterion, $16.6\%$ $[13.6, 20.1]$; on Pythia-2.8B, $64$ ($12.8\%$ $[10.2, 16.0]$). Restricted to email addresses, where the identifier is most distinctive, it is $21.3\%$ ($64/300$) at $6.9$B and $17.0\%$ ($51/300$) at $2.8$B, against raw reproduction of $26.7\%$ and $22.7\%$. For each of those documents the model reproduced \emph{that document's} identifier and not a string it emits generically. No aggregate metric and no blind attack can expose this: it is per-document, verbatim, and identity-revealing. (All identifiers are masked in reporting; no raw data is released.)

\paragraph{Concentrated in code, but prose is not spared.} Splitting the $500$ probes by whether the prefix is code, markup or configuration, document-specific leakage is about three times higher in code, and grows with capacity in both regimes: from $12.6\%$ to $35.8\%$ ($n{=}95$) in code and from $4.0\%$ to $12.1\%$ ($n{=}405$) in prose across $410$M to $6.9$B. Prose is thus the weaker regime but not a null one; at $6.9$B its interval is $[9.3, 15.6]$, well clear of zero. The aggregate rate is an average over the two, not a uniform risk.

\subsection{Extraction grows with model scale, and deduplication does not measurably reduce it}
\label{sec:ext_scale}

\begin{figure}[!htbp]
  \centering
  \includegraphics[width=\linewidth]{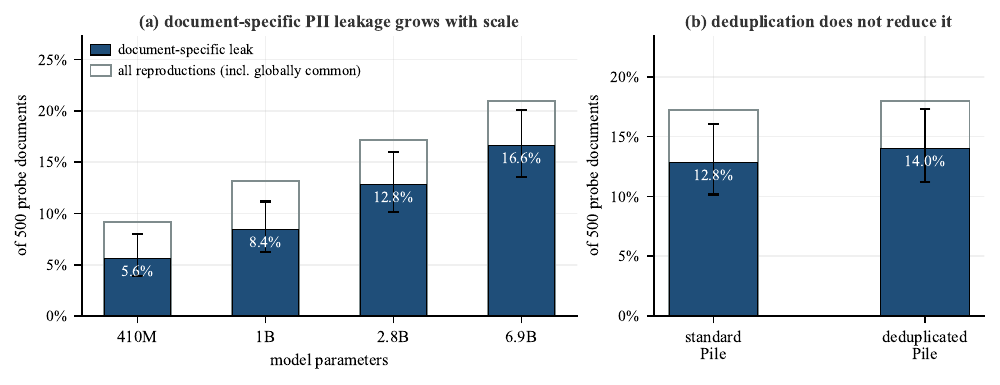}
  \caption{Verbatim extraction of real identifiers from Pile-train documents on a shared $500$-document probe set ($N{=}32$, $T{=}1.0$). Solid bars count documents whose identifier was reproduced \emph{and} was not reproduced under a mismatched prefix, so each one is attributable to that document; the outline adds the reproductions the context control discounts as globally common. Whiskers are Wilson $95\%$ intervals on the solid bars. \textbf{(a)} document-specific leakage triples from $410$M to $6.9$B while the discounted portion stays flat. \textbf{(b)} training on the deduplicated Pile does not lower it (paired McNemar $p{=}0.52$ on raw reproduction).}
  \label{fig:scale_defense}
\end{figure}

We repeat the measurement across the Pythia ladder, every model seeing the identical $500$ documents (Figure~\ref{fig:scale_defense}a). Document-specific leakage rises monotonically with capacity: $5.6\%$ $[3.9, 8.0]$ at $410$M, $8.4\%$ at $1$B, $12.8\%$ at $2.8$B and $16.6\%$ $[13.6, 20.1]$ at $6.9$B, a threefold increase with non-overlapping Wilson intervals between the ends. The context control is flat across the ladder ($5.6$--$6.4\%$ of documents), so the growth is a property of the models and not of the control. Email addresses leak about twice as often as telephone numbers at every scale ($21.3\%$ versus $9.5\%$ document-specific at $6.9$B). Larger models expose more individuals, consistent with memorisation growing with capacity \citep{carlini2022memorization} and with the capacity-dependence of individual risk reported for medical classifiers \citep{knolle2026disparate}.

We then evaluate the most commonly recommended training-time mitigation, corpus deduplication, by comparing Pythia-2.8B against Pythia-2.8B-deduped on the same probes (Figure~\ref{fig:scale_defense}b). Document-specific leakage is $12.8\%$ for the standard model and $14.0\%$ for the deduplicated one (raw reproduction $17.2\%$ versus $18.0\%$), and a paired McNemar test over the $500$ documents finds no significant difference ($\chi^2 {=} 0.41$, $p {=} 0.52$; both extract $77$, standard-only $9$, deduplicated-only $13$). We therefore find no detectable reduction in verbatim PII extraction from deduplication in this suite. Two caveats keep this from being a general claim about deduplication. First, our test is powered only for moderate effects: at this base rate and $n {=} 500$ a reduction of a few percentage points would not be detected, and \citet{biderman2023emergent} report that Pythia's deduplicated models do memorise ``albeit slightly'' less overall. Second, the comparison inherits a confound of the suite: near-deduplication (MinHashLSH at threshold $0.87$) shrinks the corpus from ${\approx}300$B to $207$B tokens while both suites are trained to the same ${\approx}300$B-token budget, so the deduplicated models make ${\approx}1.5$ passes over their data \citep{biderman2023pythia}, which \citet{biderman2023emergent} suggest may be ``offsetting the benefits of deduplicated data''. The practical implication stands regardless of mechanism: a model owner who deduplicates at document level and at this threshold should not assume individual identifiers are thereby protected, which strengthens the case for a mitigation carrying a guarantee (Sec.~\ref{sec:discussion}).

\subsection{Members' continuations are recovered, in a tail}
\label{sec:ext_recovery}

\begin{figure}[!htbp]
  \centering
  \includegraphics[width=\linewidth]{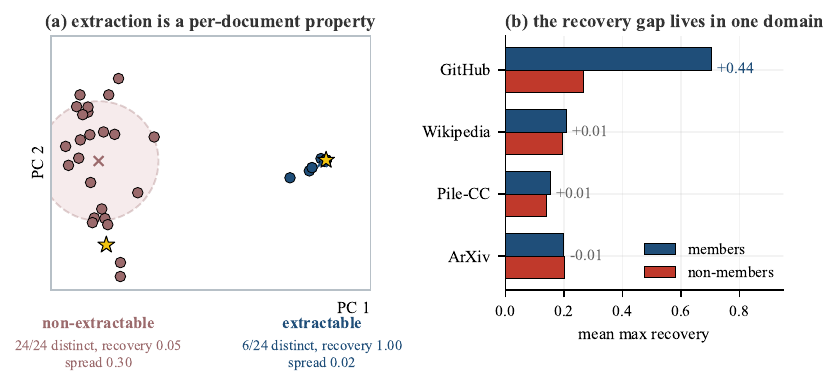}
  \caption{Why sampling reproduces some training documents and not others (Pythia-2.8B, MIMIR, $N{=}24$ shown). \textbf{(a)} two GitHub training documents in a joint PCA of their sampled continuations; each dashed circle is that document's spread, the radius enclosing $68\%$ of its samples about the centroid ($\times$). For the extractable document the samples collapse onto one another and onto the held-out training text (star), recovering it exactly, and its spread circle is smaller than a marker; for the non-extractable one every sample is distinct and the text lies outside the spread. \textbf{(b)} the pooled member/non-member recovery gap is produced almost entirely by GitHub; on prose domains it is within noise.}
  \label{fig:mechanism}
\end{figure}

The probes above target one short string per document. We now ask whether \emph{arbitrary} held-out content is recovered, which turns out to be the weaker and more domain-dependent of the two measurements. For each MIMIR document we split it into a prefix $x$ and held-out continuation $y$, sample $N$ continuations from $p_\theta(\cdot\mid x)$, and measure the maximum $5$-gram recall of $y$ across the samples (extreme-value functional). Figure~\ref{fig:mechanism}a shows the mechanism on two documents from the \emph{same} domain: for an extractable document the $24$ samples collapse onto one another ($6$ distinct strings) and onto the held-out training text, recovering it exactly; for a non-extractable document every sample is distinct and the training text lies outside the sampled cloud. Recovery of a specific held-out continuation cannot be faked by a text-only classifier and cannot occur by chance for a non-member, which is why it is informative even though its aggregate AUC ($0.57$) does not beat the blind baseline: the discriminative power lives in a high-precision tail, not in the average.

\paragraph{The recovery gap is a code phenomenon.} Pooled over MIMIR, mean max-recovery on Pythia-2.8B is $0.32$ for members versus $0.20$ for non-members, and the near-extraction tail (a sample recovering $\geq\!50\%$ of $y$) contains $18\%$ of members versus $4\%$ of non-members. That pooled number is, however, almost entirely produced by one domain (Figure~\ref{fig:mechanism}b). Per domain the member/non-member gap is $+0.44$ on GitHub ($0.71$ vs $0.27$; tail $67\%$ vs $11\%$), but only $+0.014$ on Wikipedia, $+0.013$ on Pile-CC and $-0.006$ on ArXiv. The same decomposition holds on Pythia-410M ($+0.39$ on GitHub, ${\leq}\,{+}0.05$ elsewhere). We therefore do not claim a corpus-wide recovery effect: on prose this measurement is null, and what it detects is the memorisation of \emph{code}, the most heavily duplicated content in the Pile, which is exactly where the duplication axis of \citet{carlini2022memorization} predicts memorisation to concentrate. Reporting the pooled figure alone would repeat the averaging error this paper criticises.

\subsection{The extraction regime: samples, temperature, top-$p$, and prefix length}
\label{sec:ext_regime}

Three sweeps characterise what governs extraction (Pythia-2.8B, MIMIR, recovery gap; Figure~\ref{fig:regime}, Appendix~\ref{app:regime}). \textbf{Temperature and top-$p$ have minor effect}: across $T \in \{0.3,0.7,1.0,1.3\}$ and top-$p \in \{0.9,0.95,1.0\}$ the recovery gap stays in $0.122$--$0.135$, with only a shallow edge at the temperature extremes; the identifier probes agree (Appendix~\ref{app:pii_temp}). \textbf{A short prefix already suffices}: a $16$-token prefix produces a gap of $0.13$, and longer prefixes raise absolute recovery for members and non-members alike, so the discriminative gap peaks near $32$ tokens and then narrows. \textbf{The sample budget scales inversely with prefix length}: the number of samples $N^\ast$ needed to reach $90\%$ of the $N{=}32$ gap falls from $\approx\!8$ at a $16$-token prefix to $1$ at $64$ tokens, because more context concentrates $p_\theta(\cdot\mid x)$ so a single draw suffices.

\section{Threats to Validity}
\label{sec:threats}

\paragraph{Aggregate effect size.} No sampling functional we test significantly beats a blind baseline as an \emph{aggregate} membership classifier on the IID split (Sec.~\ref{sec:res_null}), and we make no aggregate-MIA claim. Our positive result is per-document verbatim extraction (Sec.~\ref{sec:extraction}), which is not an aggregate-AUC quantity and cannot be produced by a model-free attack. The recovery \emph{gap} could still be partly driven by residual prefix-distribution differences; the extraction result (exact reproduction of a specific held-out string) is robust to this, because a non-member's identifier cannot be reproduced by chance.

\paragraph{Benchmark construction.} WikiMIA is separable from surface text alone (Sec.~\ref{sec:res_blind}) and should not be used to certify model leakage; we rely on the IID MIMIR (Pile train/test) split for all leakage claims. The Pythia-2.8B / MIMIR cell is at $77\%$ coverage ($773 / 1000$ documents); the reached source slices are balanced and reported quantities fall within the cluster-bootstrap CI of the partial run.

\paragraph{Sampling stochasticity.} Each run uses a fixed seed; completion content varies between runs due to GPU non-determinism in batched sampling. Cluster-bootstrap CIs reflect probe-level, not seed-level, variability.

\paragraph{Scope.} We study open models (Pythia, OLMo) with a known training corpus so that membership is ground-truthed. Extending the extraction measurement to closed commercial models, where the training set is unknown, is future work; the \texttt{leakit} tool already runs in that black-box setting.

\section{Discussion}
\label{sec:discussion}

\paragraph{Aggregate membership metrics are the wrong privacy lens.}
Our results and \citet{das2024blind} together show that aggregate MIA AUC on foundation models largely measures how distinguishable the member and non-member \emph{text} is, not what the model leaks. Yet the concrete privacy harm, reproducing a specific individual's training content, is real, per-document, and concentrated in a tail that aggregate metrics average away, mirroring \citet{knolle2026disparate} in the discriminative-model setting. For a model trained on a sensitive cohort, an extracted record is a direct disclosure of the sensitive attribute, not merely a re-identification risk. Privacy audits should therefore report per-document extraction (and its worst-case tail), not a single aggregate AUC.

\paragraph{Why extraction and membership diverge.}
Training sharpens $p_\theta(\cdot\mid x)$ toward the memorised continuation for a subset of documents, and sampling surfaces those continuations verbatim. This yields high-precision extraction for a tail of records while barely moving the aggregate distribution, so an averaged AUC stays near a blind baseline even as specific records leak.

\paragraph{Defenses.}
(D1) \emph{Differential privacy} bounds any single record's influence and is the only mitigation with a guarantee; \citet{knolle2026disparate} show that record-level DP can still leave a tail exposed, arguing for per-example (here, per-document) accounting, and \citet{lukas2023analyzing} quantify the resulting utility trade-off for PII specifically. (D2) \emph{Training-time deduplication} is the standard recommendation \citep{lee2022deduplicating}, but we measure no detectable reduction in identifier extraction between Pythia-2.8B and its deduplicated counterpart (Sec.~\ref{sec:ext_scale}). Whether this reflects the limits of document-level near-deduplication or the ${\approx}1.5$-epoch training budget of that suite, deduplication as instantiated there is not a substitute for a guarantee. (D3) \emph{Inference-time controls} (temperature, nucleus truncation) barely affect extraction in our sweep (Sec.~\ref{sec:ext_regime}) and are not a reliable defense; output filtering against a known-sensitive list is the exception, since extraction is verbatim and therefore detectable at the string level.

\paragraph{An audit tool.}
Because the probe needs only a sampling endpoint, a model owner can run it against their own deployment before release, under the same access a third party would have; we release \texttt{leakit} to make that routine.

\section{Conclusion}
\label{sec:conclusion}

Aggregate membership inference on black-box language models is confounded: a model-free blind baseline matches or beats every sampling functional we test on an IID benchmark, extending \citet{das2024blind} into the sampling regime. The genuine risk is per-document verbatim extraction, which aggregate metrics and blind attacks both miss, and which is highly uneven: concentrated in code, but present and capacity-dependent in prose. Privacy audits should report per-document extraction, decomposed by domain, rather than a single AUC. We release \texttt{leakit} to make such audits routine.

\section*{Reproducibility Statement}

All target models (Pythia 410M--6.9B, OLMo-1B) and benchmarks (WikiMIA, MIMIR) are public, and membership labels come from the Pile's official train/test split, so every leakage claim is ground-truthed rather than inferred. Decoding settings, prefix and target lengths, sample budgets, and seeds are given in Sec.~\ref{sec:method} and Appendix~\ref{app:models}; each perturbation and probe is deterministic given the task identifier and seed. We release \texttt{leakit}, a command-line implementation of the extraction probe that runs against any sampling endpoint, together with analysis scripts that recompute every number and figure in this paper from the shipped per-document records without GPU access\footnote{\ifanon The tool and analysis scripts are included as anonymised supplementary material with this submission. \else \url{https://github.com/victormaricato/leakit}. \fi}. The PII experiment releases only masked aggregates and per-document booleans; no identifier, prefix, or raw completion is published.

\section*{Ethics Statement}

This work measures how much training data a language model reproduces verbatim, including personally identifying information (PII), in order to quantify a privacy risk and motivate mitigations. We took the following precautions. All experiments target \emph{open} models (Pythia, OLMo) whose training corpus (the Pile) is public, so we create no new exposure: any content our probes recover is already retrievable from the published corpus. We did not attack a deployed commercial system, and we targeted no specific individual or organisation; probes were selected by a regular-expression filter over a corpus stream, not by searching for a person. Every identifier is masked at the point of measurement (\texttt{d***@e***.com}); no raw identifier appears in this paper, our logs, or the released artifact, and we publish no extracted content. We report only aggregate rates and per-document booleans. The dual-use consideration is explicit: the same procedure is an attack and an audit, which is why we release it as an audit tool and pair it with a defensive recommendation (differential privacy with per-document accounting, and training-time deduplication) rather than an extraction pipeline. We believe publication is net-beneficial because the leakage we document is already reachable by any party with sampling access, and because our central methodological finding, that aggregate membership metrics understate per-document exposure, directly affects how privacy audits should be conducted.

\section*{Use of Large Language Models}

Per the ICLR 2027 AI policy we disclose the following. A general-purpose LLM assistant was used
throughout this project, and its contribution was substantial rather than incidental. Specifically,
it was used to \emph{refine hypotheses}: the original framing of this work was a continuation-free
membership-inference attack based on self-concentration, and the reframing toward per-document
verbatim extraction followed from control experiments that showed the original thesis did not
survive a blind baseline. It was used to \emph{design experiments and provide feedback on
methodology}, including the blind-baseline comparison, the IID Pile train/test evaluation, the
mismatched-prefix context control, the per-document conjunction used as the headline statistic in
Sec.~\ref{sec:ext_pii}, the temperature, nucleus, prefix-length and sample-budget sweeps, and the
deduplication comparison. It was used to \emph{implement methods}: the sampling harness, the scorers,
the probe construction, the identifier detection and masking, the analysis scripts that recompute
every reported number, and the released \texttt{leakit} tool. It was used to \emph{clean and
reformat data} when assembling the probe sets from the Pile, and to \emph{interpret results}. We
additionally used it for literature search and for verifying that each cited claim matches its
source, for generating the figures programmatically, and for drafting and editing the prose of this
paper.

The author directed the research, chose which results to report, and is responsible for all claims
here. Every quantitative result in this paper is reproducible offline from the shipped per-document
records by the released analysis script, independently of any LLM.

\bibliographystyle{iclr2027_conference}

\begin{thebibliography}{99}\small

\bibitem[Biderman et al.(2023)]{biderman2023pythia}
Biderman, S., Schoelkopf, H., Anthony, Q., Bradley, H., O'Brien, K., Hallahan, E., et al.
\newblock {Pythia}: A suite for analyzing large language models across training and scaling.
\newblock In \emph{International Conference on Machine Learning (ICML)}, 2023.

\bibitem[Carlini et al.(2021)]{carlini2021extracting}
Carlini, N., Tramèr, F., Wallace, E., Jagielski, M., Herbert-Voss, A., Lee, K., et al.
\newblock Extracting training data from large language models.
\newblock In \emph{USENIX Security}, 2021.

\bibitem[Carlini et al.(2022)]{carlini2022memorization}
Carlini, N., Ippolito, D., Jagielski, M., Lee, K., Tramer, F., and Zhang, C.
\newblock Quantifying memorization across neural language models.
\newblock In \emph{ICLR}, 2023.

\bibitem[Das et al.(2025)]{das2024blind}
Das, D., Zhang, J., and Tramèr, F.
\newblock Blind baselines beat membership inference attacks for foundation models.
\newblock In \emph{DATA-FM Workshop at ICLR}, 2025. arXiv:2406.16201.

\bibitem[Cooper et al.(2026)]{cooper2026books}
Cooper, A.~F., Lemley, M.~A., Casasola, A., Ahmed, A., Gokaslan, A., Cyphert, A.~B., De~Sa, C., Ho, D.~E., and Liang, P.
\newblock Extracting memorized pieces of (copyrighted) books from open-weight language models.
\newblock In \emph{COLM}, 2026. arXiv:2505.12546.

\bibitem[Hayes et al.(2025a)]{hayes2025probabilistic}
Hayes, J., Swanberg, M., Chaudhari, H., Yona, I., Shumailov, I., Nasr, M., Choquette-Choo, C.~A., Lee, K., and Cooper, A.~F.
\newblock Measuring memorization in language models via probabilistic extraction.
\newblock In \emph{NAACL}, 2025. arXiv:2410.19482.

\bibitem[Hayes et al.(2025b)]{hayes2025limits}
Hayes, J., Shumailov, I., Choquette-Choo, C.~A., Jagielski, M., Kaissis, G., Nasr, M., Ghalebikesabi, S., Annamalai, M.~S.~M.~S., Mireshghallah, N., Shilov, I., Meeus, M., de~Montjoye, Y.-A., Lee, K., Boenisch, F., Dziedzic, A., and Cooper, A.~F.
\newblock Exploring the limits of strong membership inference attacks on large language models.
\newblock In \emph{NeurIPS}, 2025. arXiv:2505.18773.

\bibitem[Duan et al.(2024)]{duan2024mimir}
Duan, M., Suri, A., Mireshghallah, N., Min, S., Shi, W., Zettlemoyer, L., et al.
\newblock Do membership inference attacks work on large language models?
\newblock In \emph{COLM}, 2024.

\bibitem[Knolle et al.(2026)]{knolle2026disparate}
Knolle, M. A., Menten, M. J., Jungmann, F., Meissen, F., Glocker, B., Rueckert, D., and Kaissis, G.
\newblock Disparate privacy risks from medical {AI}.
\newblock \emph{Nature}, 2026.

\bibitem[Groeneveld et al.(2024)]{groeneveld2024olmo}
Groeneveld, D., Beltagy, I., Walsh, P., Bhagia, A., Kinney, R., Tafjord, O., et al.
\newblock {OLMo}: Accelerating the science of language models.
\newblock In \emph{ACL}, 2024.

\bibitem[Kaneko et al.(2025)]{kaneko2025samia}
Kaneko, M., Ma, Y., Wata, Y., and Okazaki, N.
\newblock Sampling-based pseudo-likelihood for membership inference attacks.
\newblock In \emph{Findings of ACL}, 2025. arXiv:2404.11262.

\bibitem[Biderman et al.(2023)]{biderman2023emergent}
Biderman, S., Prashanth, U.~S., Sutawika, L., Schoelkopf, H., Anthony, Q., Purohit, S., and Raff, E.
\newblock Emergent and predictable memorization in large language models.
\newblock In \emph{NeurIPS}, 2023. arXiv:2304.11158.

\bibitem[Lee et al.(2022)]{lee2022deduplicating}
Lee, K., Ippolito, D., Nystrom, A., Zhang, C., Eck, D., Callison-Burch, C., and Carlini, N.
\newblock Deduplicating training data makes language models better.
\newblock In \emph{ACL}, 2022. arXiv:2107.06499.

\bibitem[Lukas et al.(2023)]{lukas2023analyzing}
Lukas, N., Salem, A., Sim, R., Tople, S., Wutschitz, L., and Zanella-B\'eguelin, S.
\newblock Analyzing leakage of personally identifiable information in language models.
\newblock In \emph{IEEE Symposium on Security and Privacy (S\&P)}, 2023. arXiv:2302.00539.

\bibitem[Nasr et al.(2025)]{nasr2025scalable}
Nasr, M., Rando, J., Carlini, N., Hayase, J., Jagielski, M., Cooper, A.~F., Ippolito, D., Choquette-Choo, C.~A., et al.
\newblock Scalable extraction of training data from aligned, production language models.
\newblock In \emph{International Conference on Learning Representations (ICLR)}, 2025. arXiv:2311.17035.

\bibitem[Maini et al.(2024)]{maini2024llm}
Maini, P., Jia, H., Papernot, N., and Dziedzic, A.
\newblock {LLM} dataset inference: Did you train on my dataset?
\newblock In \emph{NeurIPS}, 2024.

\bibitem[Xie et al.(2024)]{xie2024recall}
Xie, R., Wang, J., Huang, R., Zhang, M., Ge, R., Pei, J., Gong, N. Z., and Dhingra, B.
\newblock {ReCaLL}: Membership inference via relative conditional log-likelihoods.
\newblock In \emph{EMNLP}, 2024.

\bibitem[Zaree et al.(2026)]{zaree2026attenmia}
Zaree, P., Mamun, M.~A.~A., Dong, Y., Alouani, I., and Abu-Ghazaleh, N.
\newblock {AttenMIA}: LLM membership inference attack through attention signals.
\newblock arXiv preprint arXiv:2601.18110, 2026.

\bibitem[He et al.(2025)]{he2025petal}
He, Y., et al.
\newblock Towards label-only membership inference attack against pre-trained large language models.
\newblock In \emph{USENIX Security}, 2025. arXiv:2502.18943.

\bibitem[Fu et al.(2024)]{fu2024spvmia}
Fu, W., et al.
\newblock Practical membership inference attacks against fine-tuned large language models via self-prompt calibration.
\newblock In \emph{NeurIPS}, 2024. arXiv:2311.06062.

\bibitem[Li et al.(2023)]{li2023mope}
Li, M., Wang, J., Wang, J., and Neel, S.
\newblock {MoPe}: Model perturbation-based privacy attacks on language models.
\newblock In \emph{EMNLP}, 2023.

\bibitem[Mattern et al.(2023)]{mattern2023neighborhood}
Mattern, J., Mireshghallah, F., Jin, Z., Schölkopf, B., Sachan, M., and Berg-Kirkpatrick, T.
\newblock Membership inference attacks against language models via neighbourhood comparison.
\newblock In \emph{ACL Findings}, 2023.

\bibitem[Shi et al.(2024)]{shi2024mink}
Shi, W., Ajith, A., Xia, M., Huang, Y., Liu, D., Blevins, T., et al.
\newblock Detecting pretraining data from large language models.
\newblock In \emph{ICLR}, 2024.

\bibitem[Tirumala et al.(2022)]{tirumala2022memorization}
Tirumala, K., Markosyan, A. H., Zettlemoyer, L., and Aghajanyan, A.
\newblock Memorization without overfitting: Analyzing the training dynamics of large language models.
\newblock In \emph{NeurIPS}, 2022.

\bibitem[Yeom et al.(2018)]{yeom2018loss}
Yeom, S., Giacomelli, I., Fredrikson, M., and Jha, S.
\newblock Privacy risk in machine learning: Analyzing the connection to overfitting.
\newblock In \emph{IEEE CSF}, 2018.

\bibitem[Zhang et al.(2025)]{zhang2024minkpp}
Zhang, J., Sun, J., Yeats, E., Ouyang, Y., Kuo, M., Zhang, J., Yang, H., and Li, H.
\newblock Min-K\%++: Improved baseline for detecting pre-training data from large language models.
\newblock In \emph{ICLR}, 2025.

\end{thebibliography}

\appendix
\section{Models and decoding details}
\label{app:models}

All sampling runs used $k_{\text{prefix}} {=} 64$, $\ell_{\text{target}} {=} 64$ tokens unless otherwise noted, temperature $T {=} 1.0$, top-$p {=} 1.0$, top-$k {=} 0$ (no truncation), $N {=} 32$ samples per probe. Target models are pulled from HuggingFace at their default revisions: \texttt{EleutherAI/pythia-\{410m,1b,2.8b,6.9b\}}, \texttt{allenai/OLMo-1B-hf}. The Ratio baseline uses \texttt{EleutherAI/pythia-160m} as the reference for every target (cross-family for OLMo, consistent with the Pythia-reference convention used in MIMIR's tables \citep{duan2024mimir}). Logit baselines and sampling are computed in the same forward-pass pipeline on the same probe; per-document checkpointing makes runs idempotent. All experiments ran on single commodity GPUs (T4 for $\leq 2.8$B, A10G for 6.9B), totalling ${<}50$ GPU-hours.

\section{Full scorer-family grid}
\label{app:full_grid}

Table~\ref{tab:full_grid} reports ROC-AUC for all 16 statistics defined in Sec.~\ref{sec:method:scorers} (extreme, mean-overlap, variance, self-concentration, logit baselines) across all eight (model, benchmark) configurations. Raw JSON, including TPR@5\%FPR, lives in \texttt{paper/data\_points/distribution\_stats.json}.

\begin{table}[!htbp]
\centering
\tiny
\caption{Per-configuration ROC-AUC across all evaluated statistics. Sampling-only statistics in the left block; logit baselines in the right block. Boldface marks the highest sampling-only statistic per row.}
\label{tab:full_grid}
\setlength{\tabcolsep}{2.6pt}
\renewcommand{\arraystretch}{1.05}
\begin{tabular}{ll|ccc|ccc|cc|cc|cc|cccc}
\toprule
& & \multicolumn{3}{c|}{Extreme} & \multicolumn{3}{c|}{SaMIA (mean ROUGE)} & \multicolumn{2}{c|}{Mean} & \multicolumn{2}{c|}{Variance} & \multicolumn{2}{c|}{Self-conc.} & \multicolumn{4}{c}{Logit baselines} \\
Model & Bench & 5g & LCS & Jacc & R1 & R2 & RL & 5g & Jacc & 5g & Jacc & 5g & uniq & LOSS & Zlib & MinK & Ratio \\
\midrule
P-410M & Wiki  & \textbf{0.563} & 0.488 & 0.528 & 0.561 & 0.536 & 0.534 & 0.530 & 0.532 & 0.444 & 0.459 & \textbf{0.577} & 0.503 & 0.572 & 0.622 & 0.584 & 0.591 \\
P-1B   & Wiki  & 0.554 & 0.480 & 0.500 & 0.541 & 0.542 & 0.516 & 0.520 & 0.518 & 0.468 & 0.468 & \textbf{0.575} & 0.500 & 0.591 & 0.640 & 0.601 & 0.599 \\
P-2.8B & Wiki  & \textbf{0.568} & 0.510 & 0.529 & 0.542 & 0.540 & 0.534 & 0.533 & 0.525 & 0.508 & 0.487 & 0.564 & 0.500 & 0.619 & 0.664 & 0.642 & 0.622 \\
P-6.9B & Wiki  & \textbf{0.604} & 0.498 & 0.561 & 0.574 & 0.562 & 0.546 & 0.551 & 0.547 & 0.530 & 0.509 & 0.557 & 0.500 & 0.645 & 0.690 & 0.671 & 0.675 \\
OLMo-1B & Wiki & 0.514 & 0.440 & 0.504 & \textbf{0.559} & 0.527 & 0.531 & 0.503 & 0.521 & 0.454 & 0.484 & 0.508 & 0.482 & 0.499 & 0.547 & 0.517 & 0.433 \\
\midrule
P-410M & MIMIR & \textbf{0.599} & 0.544 & 0.571 & 0.564 & 0.585 & 0.557 & 0.558 & 0.573 & 0.526 & 0.528 & 0.559 & 0.549 & 0.569 & 0.614 & 0.584 & 0.501 \\
P-1B   & MIMIR & \textbf{0.601} & 0.548 & 0.577 & 0.565 & 0.587 & 0.562 & 0.566 & 0.577 & 0.518 & 0.528 & 0.559 & 0.554 & 0.572 & 0.611 & 0.586 & 0.497 \\
P-2.8B & MIMIR & \textbf{0.614} & 0.542 & 0.571 & 0.577 & 0.591 & 0.565 & 0.565 & 0.582 & 0.478 & 0.501 & 0.559 & 0.567 & 0.571 & 0.630 & 0.601 & 0.520 \\
\midrule
\multicolumn{2}{l|}{Macro mean} & 0.577 & 0.506 & 0.543 & 0.560 & 0.559 & 0.543 & 0.541 & 0.547 & 0.491 & 0.495 & 0.557 & 0.519 & 0.580 & 0.627 & 0.598 & 0.555 \\
\bottomrule
\end{tabular}
\end{table}

\section{Recovery versus sample budget}
\label{app:recovery_n}

\begin{figure}[!htbp]
  \centering
  \includegraphics[width=0.50\linewidth]{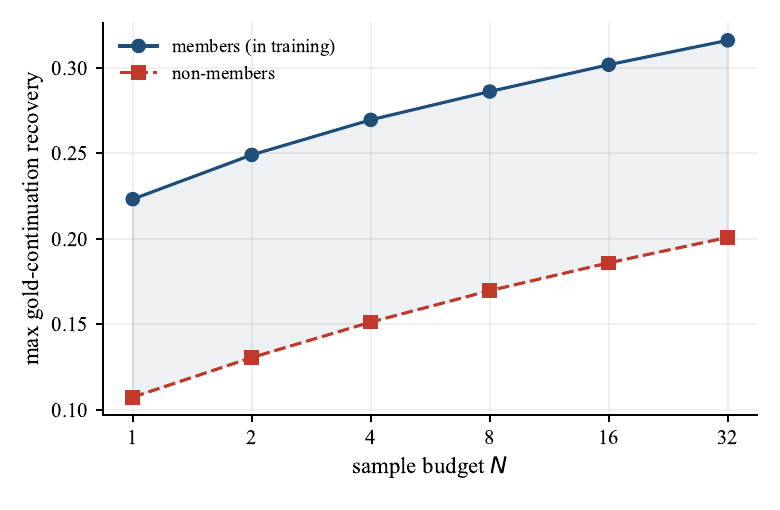}
  \caption{Max 5-gram recall of the held-out continuation versus sample budget $N$
  (Pythia-2.8B, MIMIR, pooled over domains).}
  \label{fig:recovery}
\end{figure}

Figure~\ref{fig:recovery} plots how much of a document's held-out continuation is
reproduced as the sample budget $N$ grows (Pythia-2.8B, MIMIR, pooled over domains;
see Sec.~\ref{sec:ext_recovery} for the per-domain decomposition). The member/non-member
separation is present from $N{=}1$ and more samples raise absolute recovery for both
groups. The sample budget needed to reach it, $N^\ast$, is reported per prefix length in
Sec.~\ref{sec:ext_regime}.

\section{Temperature and the PII context control}
\label{app:pii_temp}

\begin{figure}[!htbp]
  \centering
  \includegraphics[width=0.42\linewidth]{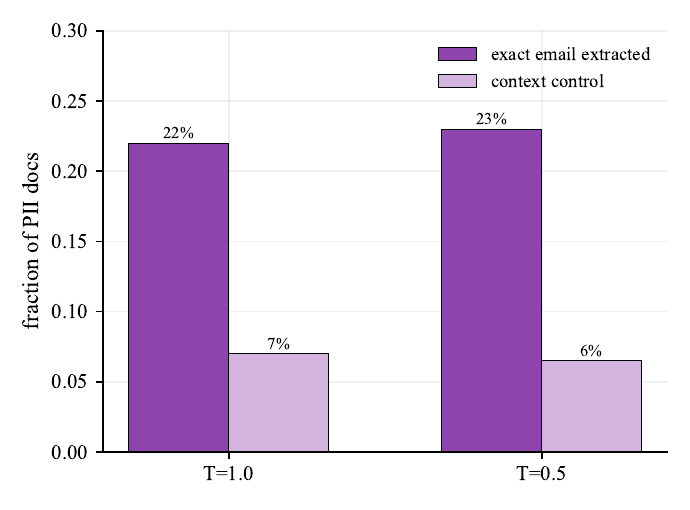}
  \caption{Verbatim email extraction versus the context control (Pythia-2.8B, $n{=}200$).}
  \label{fig:pii}
\end{figure}

Figure~\ref{fig:pii} shows the verbatim email-extraction rate against the context
control at two temperatures on the initial $200$-document probe set. The larger
$500$-document probe set used in Sec.~\ref{sec:ext_scale} supersedes it; both agree
that the control accounts for roughly a third of raw extractions.

\section{Pipeline overview}
\label{app:pipeline}

\begin{figure}[!htbp]
  \centering
  \includegraphics[width=0.82\linewidth]{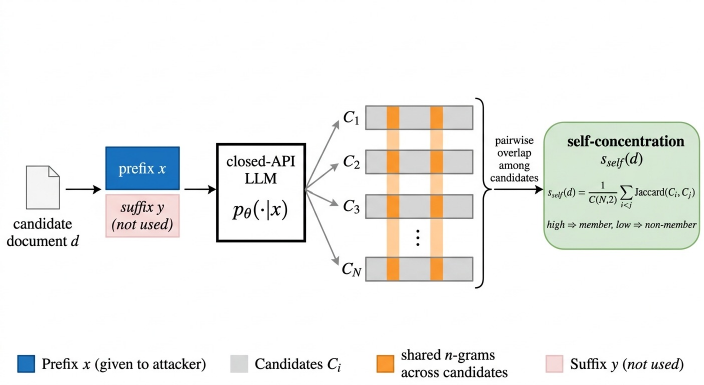}
  \caption{Probabilistic black-box leakage. From a \textbf{prefix} $x$ we query a black-box LLM for $N$ continuations $C_1, \dots, C_N \sim p_\theta(\cdot \mid x)$, an empirical estimate of the model's output distribution. Leakage signals are functionals of these samples: overlap with a candidate continuation, and the mutual overlap among the samples themselves (shown). For a memorised (member) prefix the samples concentrate on the training continuation and can reproduce it \emph{verbatim}; for a novel prefix they diverge. Orange bands mark $n$-grams that recur across the sampled set.}
  \label{fig:pipeline}
\end{figure}

\section{Extraction regime sweeps}
\label{app:regime}

\begin{figure}[!htbp]
  \centering
  \includegraphics[width=0.85\linewidth]{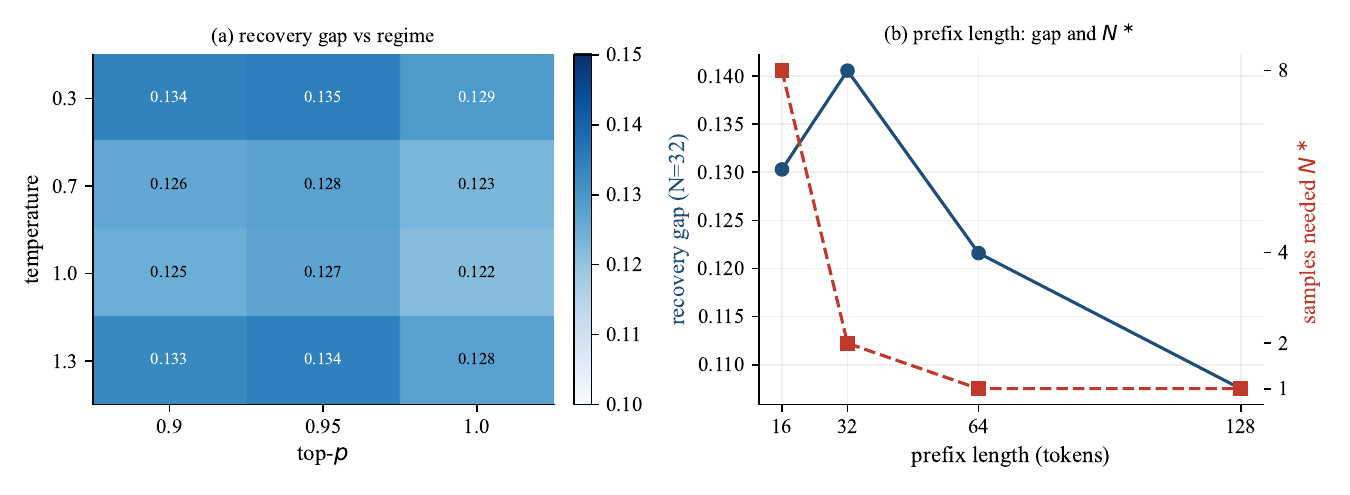}
  \caption{Extraction regime (Pythia-2.8B, MIMIR). \textbf{(a)} recovery gap across temperature and top-$p$ (minor effect). \textbf{(b)} recovery gap and the samples needed $N^\ast$ as a function of prefix length: a short prefix needs more samples, a longer prefix needs one.}
  \label{fig:regime}
\end{figure}

\section{Cost analysis}
\label{app:cost}

\paragraph{Sampling cost.} A single extraction probe at $N {=} 32$, $\ell {=} 64$ requires ${\approx}2$k generated tokens per document. At current frontier-API prices (e.g.\ \$10/M output for a frontier-class model) this is ${\sim}\$0.02$ per document, so auditing a $1000$-document corpus costs ${\sim}\$20$. The probe is well within budget for any realistic privacy-audit deployment, and a $16$-token prefix at $N{=}8$ (Sec.~\ref{sec:ext_regime}) reduces this several-fold.

\end{document}